\pdfoutput=1
\documentclass[11pt]{article}

\usepackage{EMNLP2022}
\usepackage{todonotes}
\usepackage{times}
\usepackage{latexsym}
\usepackage{array,multirow,graphicx}
\usepackage[thinlines]{easytable}
\usepackage[T1]{fontenc}
\usepackage[utf8]{inputenc}
\usepackage{multirow}
\usepackage{booktabs}
\usepackage{subfig}
\usepackage{float}
\usepackage{array}
\usepackage{ulem}

\usepackage{microtype}

\usepackage{inconsolata}

\title{Multimodal Audio-textual Architecture for Robust Spoken Language Understanding}

\author{Anderson R. Avila \\
  INRS-EMT \\
  Montréal, Québec, Canada \\
  \And
  Mehdi Rezagholizadeh \\
  Noah's Ark Lab \\
  Montréal, Québec, Canada \\
   \\\And
  Chao Xing \\
  Noah's Ark Lab \\
  Montréal, Québec, Canada \\
  }

\begin{document}
\maketitle
\begin{abstract}
Recent voice assistants are usually based on the cascade spoken language understanding (SLU) solution, which consists of an automatic speech recognition (ASR) engine and a natural language understanding (NLU) system. Because such approach relies on the ASR output, it often suffers from the so-called ASR error propagation. In this work, we investigate impacts of this ASR error propagation on state-of-the-art NLU systems based on pre-trained language models (PLM), such as BERT and RoBERTa.
Moreover, a multimodal language understanding (MLU) module is proposed to mitigate SLU performance degradation caused by errors present in the ASR transcript. The MLU benefits from self-supervised features learned from both audio and text modalities, specifically  Wav2Vec for speech and Bert/RoBERTa for language. Our MLU combines an encoder network to embed the audio signal and a text encoder to process text transcripts followed by a late fusion layer to fuse audio and text logits. We found that the proposed MLU showed to be robust towards poor quality ASR transcripts, while the performance of BERT and RoBERTa are severely compromised. Our model is evaluated on five tasks from three SLU datasets and robustness is tested using ASR transcripts from three ASR engines. Results show that the proposed approach effectively mitigates the ASR error propagation problem, surpassing the PLM models' performance across all datasets for the academic ASR engine.
\end{abstract}

\section{Introduction}

\begin{figure}
\centering
\includegraphics[width=1\linewidth]{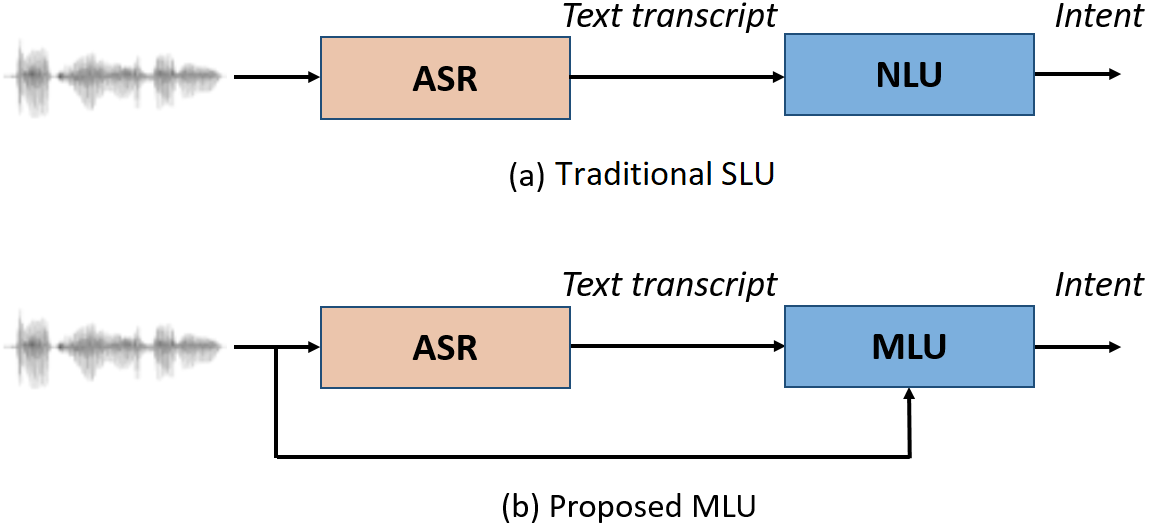}

\caption[Two numerical solutions]{Traditional SLU vs our proposed SLU architectures. The former relies solely on ASR transcripts to extract semantics whereas the latter fuses audio and text data to improve robustness of the SLU system.}
\label{fig:proposed} 
\end{figure}

Speech signals carry out the linguistic message, with speaker intentions, as well as his/her specific traits and emotions. As depicted in Figure~\ref{fig:proposed}-a, to extract semantic meaning from audio, traditional spoken language understanding (SLU) uses a pipeline that starts with an automatic speech recognizer (ASR) that transcribes the linguistic information into text, and a natural language understanding (NLU) module that interprets the ASR textual output. Such solutions offer several drawbacks \cite{serdyuk2018towards}\cite{bastianelli2020slurp}. First, the NLU relies on ASR transcripts to attain the semantic information. Because the ASR is not error-free, the NLU module needs to deal with ASR errors while extracting the semantic information \cite{simonnet2017asr}\cite{zhu2018robust}\cite{simonnet2018simulating}\cite{huang2020learning}. This is a major issue as error propagation significantly affects the overall SLU performance as shown in \cite{bastianelli2020slurp}. 

Another drawback of such approaches is that, in most cases, the two modules (ASR and NLU) are optimized independently with separate
objectives \cite{serdyuk2018towards}\cite{agrawal2020tie}. While the ASR is trained to transcribe the linguistic content, the NLU is optimized to extract the semantic information, commonly from clean text \cite{huang2020leveraging}. Hence, the traditional approach is not globally optimal for the SLU task. To overcome this issue, end-to-end SLU (e2e SLU) solutions have been proposed as an alternative to the ASR-NLU pipeline \cite{haghani2018audio}\cite{lugosch2019speech}. As pointed out in \cite{bastianelli2020slurp}, a recurrent problem of e2e SLU solutions is the scarcity of publicly available resources which leads to sub-optimal performance as well.

In this paper, we are interested in improving the robustness of traditional pipeline SLU systems. As depicted in Figure~\ref{fig:proposed}-b, this can be achieved by replacing the NLU module by the so-called multimodal language understanding (MLU) module. Such MLU-based solution combines text transcripts and their corresponding speech signals as multimodal input. Experiments show that our solution leads to SLU robustness as it mitigates performance degradation caused by low quality ASR transcripts. These transcripts are generated from three off-the-shelf ASR engines. SLU robustness is assessed on five SLU tasks from three datasets with different complexity: (1) the Fluent Speech Command (FSC) dataset \cite{lugosch2019speech}; (2) the SNIPS dataset \cite{saade2019spoken}; and (3) the recent released and challenging Spoken Language Understanding Resource Package (SLURP) dataset \cite{bastianelli2020slurp}. The contributions of this work can be summarized as follows. First, we show that state-of-the-art language models, such as BERT \cite{devlin2018bert} and RoBERTa \cite{liu2019roberta} are susceptible to the ASR error propagation problem. Second,we propose a multimodal architecture that combines speech signal and text to improve the performance of traditional SLU solutions in presence of low quality ASR text transcription.

The remainder of this document is organized as follows. In Section 2, we review the related work on SLU and multimodal approaches. Section 3 presents the proposed method. Section 4 describes our experimental setup and Section 5 discusses our results. Section 6 gives the conclusion and future works.

\section{Related Work}
% \usepackage{todonotes}
%\textbf{Joint ASR+NLU optimization}. 
One drawback of traditional SLU solutions is that the ASR and the NLU modules are optimized separately. There are different approaches in the literature to mitigate this problem. For example, in \cite{kim2017onenet}, the authors jointly train an online SLU and a language model. They show that a multi-task solution that learns to predict intent and slot labels together with the arrival of new words can achieve good performance in intent detection and language modeling with a small degradation on the slot filling task when compared to independently trained models. In \cite{haghani2018audio}, the authors propose to jointly optimize both ASR and NLU modules to improve performance. Several e2e SLU encoder-decoder architectures are explored. It is shown that an e2e SLU solution that performs domain, intent and argument predictions and is jointly trained with an e2e model to generate transcripts from the same audio input can achieve better performance. 
% It is shown that better performance is achieved when an e2e SLU solution that performs domain, intent, and argument predictions is jointly trained with an e2e model that learns to generate transcripts from the same audio input. 
This study provides two important considerations. First, joint optimization induces the model to learn from errors that matter more for SLU. Second, the authors found from their experimental results that direct prediction of semantics from audio, neglecting the ground truth transcript, leads to sub-optimal performance.

%\noindent\textbf{End-to-end SLU}. 
Recently, we  have  witnessed  an  increasing interest in minimizing SLU latency as in the joint optimization problem with e2e SLU models. Such solutions bypass the need of an ASR and extracts semantics directly from the speech signal. In \cite{lugosch2019speech}, for example, the authors introduce the FSC dataset and present a pre-training strategy for e2e SLU models. Their approach is based on using ASR targets, such as words and phonemes, that are used to pre-train the initial layers of their final model. These classifiers once trained are discarded and the embeddings from the pre-trained layers are used as features for the SLU task. The authors show that improved performance on large and small SLU training sets was achieved with the proposed pre-training approach. Similarly, in \cite{chen2018spoken}, the authors propose to fine-tune the lower layers of an end-to-end CNN-RNN based model that learns to predict graphemes. This pre-trained acoustic model is optimized with the CTC loss and then combined with a semantic model to predict intents. %A relevant and more recent research is presented in \cite{mhiri2020low}. In this work, the proposed speech-to-intent model is built based on a global max-pooling layer that allows for processing speech signals of varied length, also with the ability to process a given speech segment while receiving an upcoming segment from the same speech. In \cite{potdar2021streaming}, an end-to-end streaming SLU framework is proposed. With a unidirectional LSTM architecture, optimized with the alignment-free CTC loss, and pre-trained with the cross-entropy criterion, the authors show that their solution can predict multiple intentions in an online and incremental way. Their results are comparable to the performance of start-of-the-art non-streaming models for single-intent and multi-intent classification.

% In a more recent work \cite{cao2021sequential}, the authors propose a new architecture that combines a convolutional neural network (CNN) and an LSTM. The model is evaluated with two alignment-free losses: the connectionist temporal classification (CTC) and the connectionist temporal localization (CTL). These losses allow the streaming capability on SLU, enabling the prediction of semantics based on incoming speech. Although CTC and CTL present comparable results, one has to consider that CTL has the potential for performing localization of semantic events, whereas CTC is limited to predicting a sequence. \\

%\noindent \textbf{Multimodal SLU}. 
A well-known problem of e2e SLU solutions is their limited number of publicly available data resources (i.e. semantically annotated speech data) \cite{bastianelli2020slurp}. Because there are much more NLU resources (i.e. semantically annotated text without speech), many efforts have been made towards transfer learning techniques that enable the extraction of acoustic embeddings that borrow knowledge from state-of-the-art language models such as BERT \cite{devlin2018bert}. In \cite{huang2020leveraging}, for example, the authors propose two strategies to improve performance of e2e speech-to-intent systems with unpaired text data. The first method consists of two losses: (1) one that optimizes the entire network based on text and speech embeddings, extracted from their respective pretrained models, and are used to classify intents; and (2) another loss that minimizes the mean square error between speech and text representations. This second loss only back-propagates to the speech branch as the goal is to make speech embeddings resemble text embeddings. The second method is based on a data augmentation strategy that uses a text-to-speech (TTS) system to convert annotated text to speech. %In \cite{sari2020training}, the authors show that the performance of a speech-only e2e SLU model can be improved by training the model with non-parallel audio-textual data. For that, the authors propose a multiview learning technique based on two unimodal branches consisting of an encoder for each modality. The unimodal branches receive either text or speech as input in order to produce the output. The authors first train the text branch as more resources are available. After, the classifier is frozen and the speech encoder is trained. As the final step, both branches are fine-tuned using parallel data and the shared classifier.

While NLU systems based on pre-trained language models (PLMs) can achieve good performance on high quality transcripts, they are subject to ASR error propagation. E2e SLU systems, on the other hand, can bypass the need of an ASR but offers limited performance given the nature of the input signal and the limited amount of annotated data available. All the aforementioned works are unimodal. Although some of them combine text and speech during optimization, at inference time only one modality is used. To overcome such limitation and increase robustness of SLU systems, we propose a multimodal solution which will be introduced in the next section.

%In \cite{gu2017speech}, a multimodal deep learning architecture is proposed to extract features from audio-textual data for utterance-level speech classification. Textual and acoustic representation are attained from two separated convolutional neural network architecture and combined into a unique representation. This representation is then used by a fully-connected (FC) layer and its output is fed into a decision softmax layer.  \\

\section{Methodology}

\begin{figure}
\centering
  \includegraphics[width=0.94\linewidth]{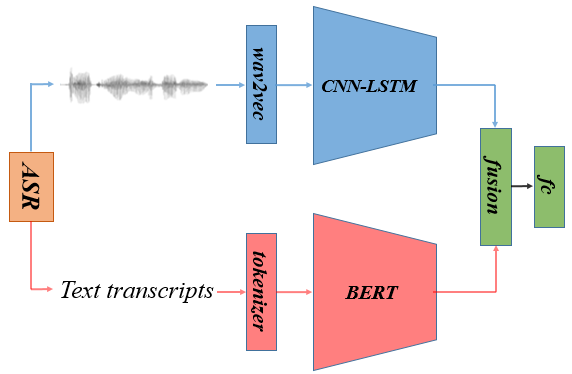}
  \caption{Diagram depicting the proposed multimodal language understanding (MLU) architecture used to predict semantic labels from audio-textual data.}
  \label{fig:architecture}
\end{figure}

In this section, we start formally describing our task. We then present the proposed architecture and finalize introducing two strategies for performing the fusion of multimodal features.

\subsection{General Principles}

As a special case of SLU, spoken utterance classification (SUC) aims at classifying the observed utterance into one of the predefined semantic classes $L = \{l_1,...,l_k\}$  \cite{masumura2018neural}. Thus, a semantic classifier is trained to maximize the class-posterior probability for a given observation, $W = \{w_1, w_2,..., w_j\}$, representing a sequence of tokens. This is achieved by the following probability:

\begin{equation}
\label{eq:suc_1}
    L^* = \arg\max_{L} P(L|W, \theta)
\end{equation}

\noindent where $\theta$ represents the parameters of the e2e neural network model. In this work, our assumption is that the robustness of such network can be improved if an additional modality, $X = \{x_1, x_2,..., x_n\}$, representing acoustic features, is combined with the text transcript. Thus, Eq. (\ref{eq:suc_1}) can be re-written as follow:

\begin{equation}
\label{eq:su_2}
    L^* = \arg\max_{L} P(L|W, X, \theta)
\end{equation}

\subsection{Architecture Overview}
The proposed architecture consists of a speech encoder based on the pre-trained speech model, wav2vec \cite{schneider2019wav2vec}, a convolutional module and a LSTM layer. As shown in Figure~\ref{fig:architecture}, the convolutional module and LSTM layer receive wav2vec embedded features as input and fine-tunes the speech representation for the downstream SLU task. This is referred to as our e2e SLU.

The text encoder, on the other hand, is based on the pretrained BERT. The encoders are trained separately on the downstream task. After optimizing each model, late fusion is adopted to combine the two modalities.

%speech embeddings with text embeddings by projecting the former into the text space. The output of the cross-modal layer is then fused with the text embeddings and presented to another LSTM network. Next, we present more details regarding our proposed architecture. We start presenting the Wav2vec speech embeddings, then we describe our speech encoder, followed by our text encoder and we finish discussing our multi-modal LSTM.
%The cross-modal layer aligns speech and text by projecting the encoded speech features onto the same space of the encoded text features. Our results show that the benefit of such approach is two fold.

\subsection{Wav2vec Embeddings}

We use the wav2vec model to extract deep semantic features from speech. While state-of-the-art models require massive amount of transcribed audio data to achieve optimal performance in speech processing tasks, wav2vec is an self-supervised pre-trained model trained on a large amount of unlabelled audio \cite{schneider2019wav2vec}. The motivation to adopt wav2vec relies on the fact that the model is able to provide a general and powerful audio representation that helps to leverage the performance of downstream tasks \cite{baevski2020wav2vec}. Thus, given an audio signal, \textbf{$x_i$} $\in$ $\mathcal{X}$, a five-layer convolutional neural network, $f : \mathcal{X} \rightarrow \mathcal{Z}$, is applied in order to obtain a low frequency feature representation, \textbf{$z_i$} $\in$ $\mathcal{Z}$, which encodes about 30 ms of audio at every 10 ms. Following, a context network, $g : \mathcal{Z} \rightarrow \mathcal{C}$, is applied to the encoded audio and adjacent embeddings, \textbf{$z_i, ..., z_v$}, are used to attain a single contextualized vector, $c_i = g(\textbf{$z_i, ..., z_v$})$. A causal convolution of 512 channels is applied to the encoder and context networks and normalization is performed across the feature and temporal dimensions for each sample. Note that $c_i$ represents roughly 210 ms of audio context with each step $i$ comprising a 512-dimensional feature vector \cite{baevski2020wav2vec}. 

%Wav2vec is based on a convolutional neural network (CNN) optimized with a contrastive loss that distinguishes between true and negative future audio samples. The model receives raw audio and computes a representation used as input to our LSTM speech encoder which we shall discuss next.

\subsection{Convolutional LSTM Speech Encoder}
\label{sec:sp_encoder}

\begin{figure}
\centering
  \includegraphics[width=0.94\linewidth]{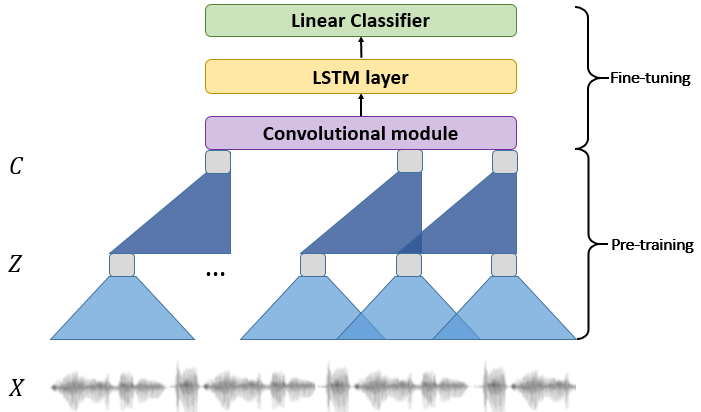}
  \caption{Models architecture combines the pre-trained wav2vec with a convolutional and lstm layers and a linear classifier.}
  \label{fig:speech_encoder}
\end{figure}

In order to fine-tune the pre-trained wav2vec for the downstream task, a convolutional module and a LSTM layer is added on top of the context network, followed by a linear classifier that projects the hidden states from the LSTM into a set of $L$ semantic labels. The architecture is depicted in Figure~\ref{fig:speech_encoder}. Our convolution module is inspired in \cite{gulati2020conformer} and consists of a gating mechanism, a point-wise convolution and a gated linear unit (GLU), which is followed by a single 1-D depthwise convolution layer. Batchnorm is deployed just after the convolution to aid training deep models. A  single-layer LSTM is also used to further improve the speech representation and was found to be relevant for the downstream SLU task. The feature dimension in the LSTM layer is controlled with a projection layer as shown bellow:

\begin{equation}
    \mathbf{s}_i = LSTM(\mathbf{c}_i), i \in  \{1...N\}
\end{equation}

\begin{equation}
    \label{eq:projection}
    \mathbf{\overline{s}}_i = W_{sp}\mathbf{s}_i
\end{equation}
\\
\noindent where $\mathbf{c}_i$ is the sequence of 512-dimensional feature representation from the convolutional layer, with $i$ being the frame index. The hidden states of the unidirectional LSTM is represented by $\mathbf{s}_i$ which is a 1024-dimensional representation that undergoes a projection layer, $W_{sp}$, leading to $\mathbf{\overline{s}}_i$. The projection layer is an alternative LSTM architecture, proposed in \cite{sak2014long}, that minimizes the computational complexity of LSTM models. In our architecture, we project a 1024-dimensional features to half of this dimension. %Thus, during the fine-tuning phase the speech encoder is optimized to output semantic labels using wav2vec embeddings as input.

\subsection{Late Score Fusion}

In order to classify semantic labels using both audio and text information, we aggregate the output probabilities given by each modality for each class. Thus, multimodal predictions are attained based on the class with the highest averaged confidence. To achieve this, we first fine-tuned the speech encoder described in Section~\ref{sec:sp_encoder} and the BERT$_{large}$ model separately. We investigated two strategies. The first one, referred to as MLU$_{avg}$, is the softmax of the avaraged probabilities as described below:

\begin{equation}
    p_l = \frac{e^{\overline{o}_l}}{\sum_{k=1}^L e^{\overline{o}_k}}
\end{equation}

\noindent where $\overline{o}$ is the averaged probability for each class. In the second method, probability aggregation \cite{cai2016simple} was used before applying the softmax.

%We also attempted different aggregation approaches based on choosing the prediction with the highest confidence.

%combining the outputs after classification. This process predicts the final output by considering the individual labels (hard level) or scores (soft level) of the involved classifiers. The following decision rules were used: majority vote (most represented class label), maximum (class label with the highest confidence), and average (class label with the highest averaged confidence).

\section{Experimental Setup}

In this section, the datasets used in our experiments and the ASR engines adopted to investigate the impact of ASR error propagation on SLU are presented. We then discuss our data augmentation strategy based on noise injection, followed by the experimental settings description.

\subsection{Datasets}

Three SLU datasets are used in our experiments. The reader is referred to Table~\ref{tbl:stats} for partial statistics covering number of speakers, number of audio files, duration (in seconds), and utterance average length (in seconds). The first is the FSC dataset which comprises single-channel audio clips sampled at 16 kHz. The data was collected using crowdsourcing, with participants requested to cite random phrases for each intent twice. It contains about 19 hours of speech, providing a total of 30,043 utterances cited by 97 different speakers. The data is split in such a way that the training set contains 14.7 hours of data, totaling 23,132 utterances from 77 speakers. Validation and test sets comprise 1.9 and 2.4 hours of speech, leading to 3,118 utterances from 10 speakers and 3,793 utterances from other 10 speakers, respectively. The dataset has a total of 31 unique intent labels resulted in a combination of three slots per audio: action, object, and location. The latter can be either “none”, “kitchen”, “bedroom”, “washroom”, “English”, “Chinese”, “Korean”, or “German”. More details about the dataset can be found in \cite{lugosch2019speech}.

\begin{figure}
\centering
  \includegraphics[width=0.99\linewidth]{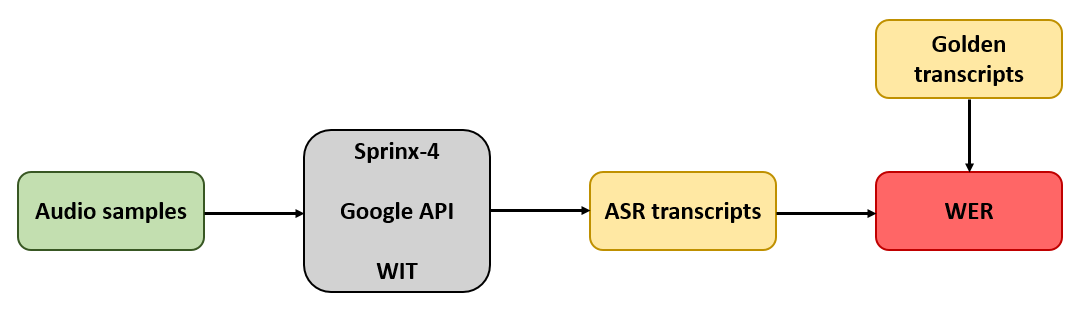}
  \caption{Pipeline for generating ASR text transcripts.}
  \label{fig:audio_asr_wer}
\end{figure}

SNIPS is the second dataset considered here. It contains a few thousand text queries. Recordings were crowdsourced and one spoken utterance was collected for each text query in the dataset. There are two domains available: smartlights (English) and smartspeakers (English and French). In our experiments only the former was used as it comprised only English sentences. With a reduced vocabulary size of approximately 400 words, the data contains 6 intents allowing to turn on or off the light, or change its brightness or color \cite{saade2019spoken}.

The recent released SLURP dataset is also considered in our experiments. It is a multi-domain dataset for end-to-end SLU and comprises approximately 72,000 audio recordings (58 hours of acoustic material), consisting of user interactions with a home assistant. The data is annotated with three levels of semantics:
Scenario, Action and Intent, having 18, 56 and 101 classes, respectively. The dataset collection was performed by first annotating textual data, which was then used as golden transcripts for audio data collection. 100 participants were asked to read out the collected prompts. This was performed in a typical home or office environment. Although SLURP offers distant and close-talk recordings, only the latter were used in our experiments. The reader is refer to \cite{bastianelli2020slurp} for more details on the dataset.

\begin{figure}
\centering
  \includegraphics[width=0.99\linewidth]{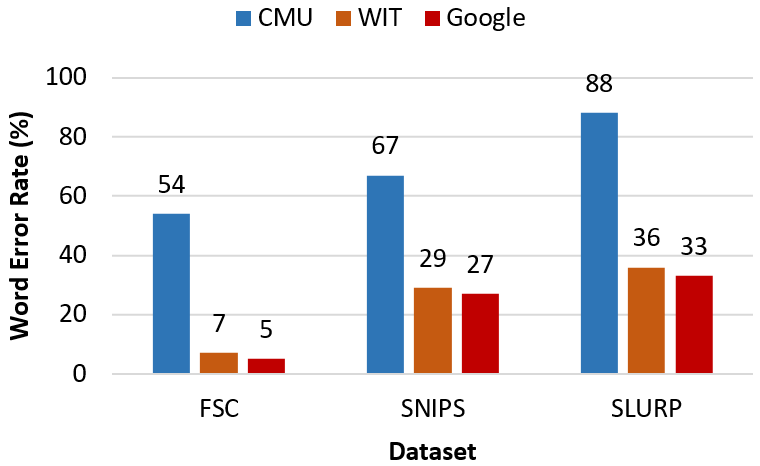}
  \caption{Word error rate (WER) based on true ASR engines (cmu, google, cloud and wit) for the three investigated datasets.}
  \label{fig:wer}
\end{figure}

\begin{table}[htb]
\centering
\scalebox{0.8}{
\begin{tabular}{cccc}
\hline
 & \textbf{FSC} & \textbf{SNIPS} & \textbf{SLURP} \\
\hline 
\textbf{$\#$ Speakers} & 97 & 69 & 177  \\
\textbf{$\#$ Audio files (headset)} & 30,043 & 2,943 & 34,603  \\
\textbf{$\#$ Audio files (Close-talk)} & - & 2,943 & 37,674  \\
\textbf{Duration [hs]} & 19 & 5.5 & 58 \\
\textbf{Avg. length [s]} & 2.3 & 3.4 & 2.9  \\
\hline
\end{tabular}}

\caption{ Statistics of audio samples for SLURP, SNIPS and FSC \cite{bastianelli2020slurp}.
}
\label{tbl:stats}
\end{table}

\begin{table*}
\centering
\scalebox{0.8}{
\begin{tabular}{llccccccccccccccc}
\hline
  & & \multicolumn{2}{c}{\textbf{FSC-I}} & & \multicolumn{2}{c}{\textbf{SNIPS-I}} & & \multicolumn{2}{c}{\textbf{SLURP-S}} & & \multicolumn{2}{c}{\textbf{SLURP-A}} & & \multicolumn{2}{c}{\textbf{SLURP-I}} \\
  \cmidrule(lr){3-4}\cmidrule(lr){6-7}\cmidrule(lr){9-10}\cmidrule(lr){12-13}\cmidrule(lr){15-16}
\textbf{Model} & \textbf{Modality} & \textbf{Acc} & \textbf{F1} & & \textbf{Acc} & \textbf{F1} & & \textbf{Acc} & \textbf{F1}& & \textbf{Acc} & \textbf{F1}& & \textbf{Acc} & \textbf{F1}\\
\hline
\textbf{e2e SLU}  & \textbf{S} &95.20 &95.21& &63.54 &63.41& & 63.88 &63.88& & 57.28 &56.77& & 50.28 & 50.05 \\\hline
\textbf{BERT} & \textbf{T} & 99.99 &100.00&  & 98.26 &98.26& & 91.98 &92.07& & 90.24 &90.19& & 86.59 & 86.38 \\%\vspace{0.25cm}
\textbf{RoBERTa} & \textbf{T} & 99.99 &100.00&  & 98.26 &98.26& & 92.76 &92.67& & 91.27 &91.22& & 86.59 & 86.38\\\hline
\textbf{MLU}$_{avg}$ & \textbf{S+T} & 99.97 &99.97&  &94.09 &94.10& & 89.95 &89.75& & 89.95 &89.75& & 84.61 & 83.92 \\%\vspace{0.25cm}
\textbf{MLU}$_{avg}$ & \textbf{S+T} & 99.99 &100.00&  & 94.79 &94.82& & 90.91 &90.80& & 90.91 &90.80& & 85.42 & 84.78\\%\vspace{0.25cm}
\hline
\end{tabular}}
\caption{
Accuracy results for the SLURP, FSC and SNIPS datasets when gold transcripts are available for training and testing the NLU, MLU and the MLU with the attention mechanism.
}
\label{tbl:mlu}
\end{table*}

Note that compared to other datasets, SLURP is much more challenging. The authors in \cite{bastianelli2020slurp}, directly compared SLURP to FSC and SNIPS in different aspects. For instance, SLURP contains 6x more sentences than SNIPS and 2.5x more audio samples than FSC. It also covers 9 times more domains and is 10 times lexically richer than both FSC and SNIPS. SLURP also provides a larger number of speakers compared to FSC and SNIPS. Next, we describe three ASR engines used to generate text transcripts. We also present the performance of these engines in terms of WER for each SLU dataset.

\subsection{ASR engines} 
\label{sec:engines}

In order to evaluate the performance of our model in a more realistic setting, we simulate the generation of text transcripts from ASR engines as depicted in Figure~\ref{fig:audio_asr_wer}. This is particularly important to assess the robustness of SLU models when golden transcripts are not available (i.e. at testing time). The ASR systems adopted here are the open-source CMU SPHINX \cite{kepuska2017comparing}, developed at Carnegie Mellon University (CMU); the Google ASR API, which enables speech to text conversion in over 120 languages \cite{google}; and the WIT engine \cite{wit}, which is an online software platform that enables the development of natural language interfaces with support to more than 130 languages. 

We evaluated the performance of these three ASR engines in terms of word error rate (WER) and the results are presented in Figure~\ref{fig:wer}. As expected, the SLURP revealed to be the most challenging dataset with the highest WER for all the three engines, followed by SNIPS and FSC. Note that, We chose datasets with different levels of complexity as well as ASR engines with diverse performance in order to evaluate our proposed MLU.

\subsection{Experimental Settings}

Our network is trained on mini-batches of 16 samples over a total of 200 epochs. Early-stopping is used in order to avoid overffiting, thus training is interrupted if the accuracy on the validation set is not improved after 20 epochs. Our model is trained using the Adam optimizer \cite{kingma2014adam}, with the initial learning rate set to $0.0001$ and a cosine learning rate schedule \cite{loshchilov2016sgdr}. Dropout probability was set to $0.3$ and the parameter for weight decay was set to $0.002$. Datasets are separated into training, validation and test sets and the hyperparameters are selected based on the performance on the validation set. All reported results are based on f1 score on the test set. 

Our experiments are based on 5 models: two NLU baselines based on BERT$_{large}$ and RoBERTa$_{large}$; an e2e SLU; and two MLU proposed solutions, MLU$_{avg}$ and MLU$_{ft}$. These models are trained to predict semantic labels for 5 tasks referred to as: FSC-I, SNIPS-I, SLURP-S, SLURP-A and SLURP-I. SLURP-S and SLURP-A denote scenario and action classification, respectively, and the remainder refer to intent classification.

\section{Experimental Results}

In this section, we present our experimental results. We start comparing the performance of the 5 aforementioned models in presence of golden transcripts. We then discuss the effects of ASR error propagation on the NLU baselines. Finally, we present the benefits of combining speech and text to overcome ASR transcript errors. 
\begin{table*}
\centering
\scalebox{0.65}{
\begin{tabular}{llcccccccccccccc}
\hline
&& \multicolumn{2}{c}{\textbf{20 \%}} & & \multicolumn{2}{c}{\textbf{40 \%}} & & \multicolumn{2}{c}{\textbf{60 \%}} & & \multicolumn{2}{c}{\textbf{80 \%}} & & \multicolumn{2}{c}{\textbf{100 \%}} \\
\cmidrule(lr){3-4}\cmidrule(lr){6-7}\cmidrule(lr){9-10}\cmidrule(lr){12-13}\cmidrule(lr){15-16}
\textbf{Task} & \textbf{Engine}& \textbf{BERT} & \textbf{RoBERTa} & & \textbf{BERT} & \textbf{RoBERTa} & & \textbf{BERT} & \textbf{RoBERTa} & & \textbf{BERT} & \textbf{RoBERTa} & & \textbf{BERT} & \textbf{RoBERTa} \\\hline
\multirow{3}{*}{\textbf{FSC-I}}&\textbf{CMU} &  89.53 & 89.74 & & 79.24  & 80.43 & & 69.51 & 71.12 & & 58.04 & 60.54 & & 50.12 & 52.67 \\
&\textbf{WIT} &  99.02 & 98.89 & & 97.70  & 97.49 & & 96.43 & 96.32 & & 95.79 & 95.35 & & 94.92 & 94.35\\
&\textbf{Google} &  99.24 & 99.29 & & 98.53  & 98.63 & & 97.89 & 98.01 & & 97.10 & 97.17 & & 96.47 & 96.65 \\\hline
\multirow{3}{*}{\textbf{SNIPS-I}}&\textbf{CMU} &  88.87 & 89.32 & & 79.18  & 80.11 & & 71.29 & 72.16 & & 60.62 & 61.41 & & 53.43 & 56.21 \\
&\textbf{WIT} &  97.22 & 96.52 & & 95.15  & 95.82 & & 93.07 & 94.48 & & 91.33 & 91.46 & & 89.27 & 90.04\\
&\textbf{Google} &  97.91 & 96.88 & & 96.53  & 95.15 & & 94.82 & 95.14 & & 93.75 & 91.73 & & 93.05 & 90.68\\\hline
\multirow{3}{*}{\textbf{SLURP-S}}&\textbf{CMU} &  81.85 & 81.92 & & 71.31  & 71.73 & & 59.68 & 60.19 & & 48.77 & 49.73 & & 38.70 & 40.32 \\
&\textbf{WIT} &  90.26 & 90.97 & & 88.65  & 89.09 & & 87.06 & 87.73 & & 85.43 & 86.08 & & 83.96 & 84.63\\
&\textbf{Google} &  90.31 & 90.77 & & 88.74  & 89.32 & & 87.07 & 87.60 & & 85.70 & 86.46 & & 84.51 & 85.01\\\hline
\multirow{3}{*}{\textbf{SLURP-A}}&\textbf{CMU} &  80.02 & 80.56 & & 69.53  & 69.79 & & 58.02 & 59.03 & & 46.06 & 47.03 & & 36.05 & 37.27 \\
&\textbf{WIT} &  88.02 & 89.17 & & 86.06 & 86.84 & & 83.00 & 84.64 & & 81.33 & 82.75 & & 79.70 & 80.99\\
&\textbf{Google} &  87.82 & 88.67 & & 85.92  & 86.96 & & 83.28 & 84.37 & & 81.71 & 82.83 & & 79.81 & 81.05\\\hline
\multirow{3}{*}{\textbf{SLURP-I}}&\textbf{CMU} &  76.14 & 76.66 & & 64.82  & 65.34 & & 52.99 & 53.22 & & 41.69 & 41.95 & & 30.91 & 31.43 \\
&\textbf{WIT} &  84.48 & 84.88 & & 82.33  & 82.72 & & 80.54 & 80.78 & & 78.57 & 79.01 & & 77.06 & 77.57\\
&\textbf{Google} &  84.14 & 84.65 & & 82.58  & 82.91 & & 80.54 & 80.72 & & 78.91 & 79.12 & & 77.52 & 77.98\\
\hline
\end{tabular}}
\caption{Effect of mixing golden transcripts with varying amount of ASR transcript output on two NLU baselines. We investigate the performance on the SLURP, FSC and SNIPS datasets using three ASR engines: CMU, WIT and Google. Performance is reported in terms of f1 score.
}
\label{tbl:effect}
\end{table*}

\subsection{Performance With Golden Transcripts}

In Table~\ref{tbl:mlu}, we present the performance of the NLU baselines, the e2e SLU and the two MLU approaches. Note that golden transcripts are available during training and testing time for the NLU and MLU systems. The former uses text-only inputs while the latter combines speech and text as input. The e2e SLU relies on speech-only input. Performance is compared in terms of accuracy and f1 scores. Across all datasets, the e2e SLU approach provides the lowest accuracy compared to the NLU and MLU solutions. This is due to the fact that models based on speech are harder to train as speech signals present more variability compared to text signals. For example, it contains the linguistic content, intra- and inter-speaker variability \cite{bent2017representation}, as well as information from the acoustic ambience. The FSC-I showed to be the easiest task with accuracy and f1 scores as high as 100 \% for all modalities, with a slight decay for the e2e SLU, which achieves roughly 95.20 \% in terms of accuracy and f1 scores. The gap between the e2e SLU performance and the other solutions is more significant for the SNIPS and SLURP tasks. For instance, BERT and RoBERTa are able to achieve 98.26 \% accuracy and f1 scores for intent classification on the SNIPS dataset while e2e SLU model achieves only 63.54 \% and 63.41, respectively. Similar trend is observed for the SLURP tasks. Note that the MLU$_{ft}$ provides better performance when compared to the MLU$_{avg}$. One explanation is that the speech features are noisier (comprising much more variability as discussed above), the fine-tuning approach tends to rely more on text rather than on complementary information from the speech signal. These results show that, when golden transcripts are available, BERT and RoBERTa will provide optimal performance compared to the e2e SLU and the MLU proposed in this work. %Results also show that the MLU will not compromise the performance, providing slight decay in terms of accuracy and f1 score, specially for the datasets with more hours of training data, such as the FSC and SLURP. 

\subsection{Impact of ASR Error on NLU Baselines}

In Table~\ref{tbl:effect}, we investigate the impact of ASR error propagation into the NLU baselines, BERT and RoBERTa. For this, transcripts sampled from CMU, WIT and Google ASR engines were mixed with golden transcript samples. This was performed only for the test set in order to emulate more realistic scenarios (i.e., beyond laboratory settings). We assume that golden transcripts will be available only at training time. We observe a similar trend across all three datasets and five tasks. Performance decays as the number of ASR transcript samples increases. The performance on the FSC dataset is the least affected by ASR outputs. This is due to the fact that the FSC is a much less challenging dataset compared to SNIPS and SLURP, as discussed in \cite{bastianelli2020slurp} and also shown in Figure~\ref{fig:wer} in terms of WER. Comparing the performance of BERT and RoBERTa when golden transcripts are available (see Table~\ref{tbl:mlu}) and when 100 \% of transcripts are from the ASR engines, we observe a decay of roughly 50 \% for the academic ASR (i.e. CMU)and 3 \% when using the two commercial ASR engines (i.e. Google and WIT). The NLU performance is also evaluated on the SNIPS-I task. We notice lower f1 score compared to the FSC-I,  which is due to the characteristic of SNIPS. It has less samples available to train the model and overall a more challenging dataset as observed in Figure~\ref{fig:wer}. The performance on the SLURP dataset is the most affected by noisy ASR transcripts. For the academic ASR engine, for example, performance in terms of f1 scores can get as low as 30.91 \%, for the SLURP-I task, and as low as 37.27 \% and 40.32 \% for SLURP-S and SLURP-A tasks, respectively. When compared to the performance attained with golden transcripts, this represents a decay of 65 \%, 59 \% and 56 \%, respectively. As shown in Figure~\ref{fig:audio_asr_wer} and discussed in \cite{bastianelli2020slurp}, SLURP is a more challenging SLU dataset. For the other two comercial ASR engines, the impact of ASR transcripts are much lower but still exists for the SLURP dataset, representing a decay in terms of accuracy of roughly 15 \%, 11 \% and 12 \% for the SLURP-I, ALURP-S and SLURP-A tasks, respectively.

\subsection{SLU Robustness Towards ASR Error Propagation}

\begin{figure} 
\begin{center}

\includegraphics[width=7.8cm]{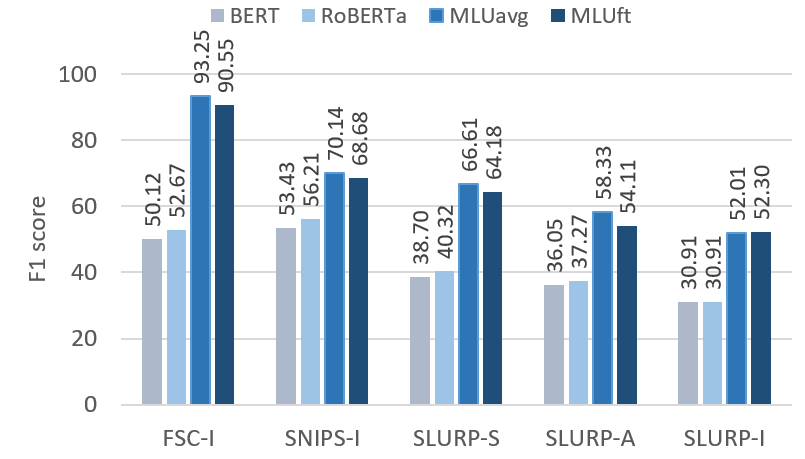}
   
\caption{SLU performance in terms of f1 score when transcripts from the CMU engine is used during test.}%
    \label{fig:cmu_error}%
    \end{center}
\end{figure}

\begin{figure} 
\begin{center}

\includegraphics[width=7.8cm]{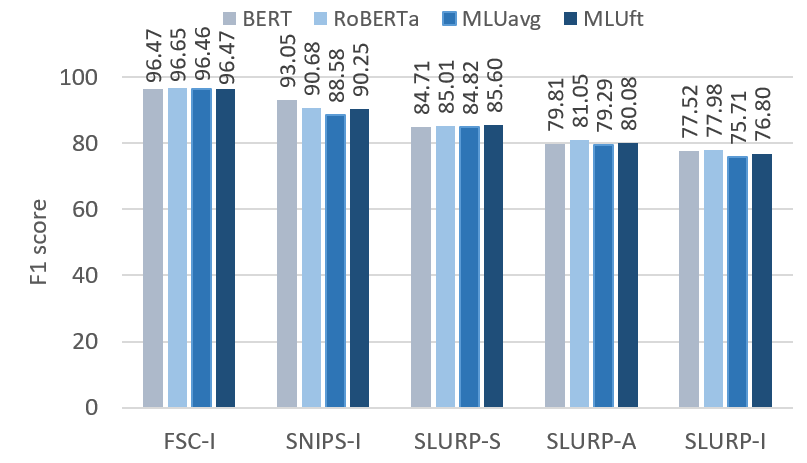}
   
\caption{SLU performance  in terms of f1 score when transcripts from the google engine is used during test.}%
    \label{fig:google_error}%
    \end{center}
\end{figure}

In this section, we evaluate the robustness of the proposed MLU towards ASR error generated by the academic ASR engine, CMU, and by the commercial engine from Google. The results are presented respectively on Figures~\ref{fig:cmu_error} and~\ref{fig:google_error}. As the commercial ASR engines have similar performance, we only present results from one of them. To evaluate a more realistic scenario, we assume no access to the golden transcripts at testing time. For all tasks, our proposed MLU model was successful towards mitigating the impact of low quality ASR transcripts attained from the academic ASR (i.e. CMU engine). We can observe that the MLU$_{avg}$ provides better performance than the MLU$_{ft}$. We hypothesize that this is because fine-tuning the model tends to rely more on the text information which is clean during the fine-tuning process. However, as we consider ASR transcripts at testing time, the text data is not as reliable as it was during training. For the commercial ASR engine, which provide higher quality ASR transcripts, performance of the proposed MLU is equivalent to the NLU baselines, showing that it can be an alternative solution to mitigate the ASR error propagation without compromising performance when text transcripts are attained with high quality. 

\subsection{Effect of Noise Injection}

Introducing noise into a neural network input is a form of data augmentation that improves robustness and leads to better generalization \cite{coulombe2018text}. To increase the robustness of our proposed model, we injected noise into the training set. We used lexical replacement which consists of proposing one or more words that can replace a given word. Thus, we choose a random word from the vocabulary $V$ with the main constraint to not be the target word $w$ in an utterance. This was achieved by perturbing golden transcripts by adding, dropping, or replacing a few words in a sentence. During training, we randomly selected 30 \% of sentences within a batch to be corrupted with noise. Moreover, only 1/3 of words within a sentence were corrupted. Table~\ref{tbl:noise_injection} presents results without and with noise injection for the NLU system based on the PLM RoBERTa and the MLU based on logits average. Results are based on low quality ASR transcripts. Noise injection was found to be beneficial for both systems (i.e. NLU and MLU), thus helping to increase robustness. For the NLU, results were more significant the FSC and SNIPS, with moderate gains for the 3 tasks presented in the SLURP. Combining noise injection with the MLU showed to be the best scenario towards mitigating the impact of ASR error propagation.%, improving f1 score across all 5 tasks. 

\begin{table}[htb]
\centering
\scalebox{0.61}{
\begin{tabular}{ccccccc}
\hline
 & & \textbf{FSC} & \textbf{SNIPS} & \textbf{SLURP-S} & \textbf{SLURP-A} & \textbf{SLURP-I} \\[1pt]
\hline 
\multirow{2}{*}{\rotatebox[origin=c]{90}{\textbf{NLU}}} &\textbf{W/o noise inj.} & 50.12 & 56.21 & \textbf{40.32} & 37.27 &  30.91 \\
&\textbf{Noise inj.} & \textbf{59.21} & \textbf{58.99} & 40.27 & \textbf{37.49} & \textbf{31.70} \\[1pt]
\hline\hline
\multirow{2}{*}{\rotatebox[origin=c]{90}{\textbf{ MLU }}} &\textbf{W/o noise inj.} & 93.25 & 72.04 & 66.61 & 58.33 & 52.01\\
&\textbf{Noise inj.} & \textbf{95.12} & \textbf{75.85} & \textbf{68.47} & \textbf{60.01} &  \textbf{54.23} \\[2pt]
\hline
\end{tabular}}

\caption{Results in terms of f1 score with noise injection and without noise injection during training.
}
\label{tbl:noise_injection}
\end{table}

\section{Conclusion}

In this paper, we propose a multimodal language understanding (MLU) architecture, which combines speech and text to predict semantic information. Our main goal is to  mitigate ASR error propagation into traditional NLU. The proposed model combines an encoder network to embed audio signals and the state-of-the-art BERT to process text transcripts. Two fusion approaches are explored and compared. A pooling average of probabilities from each modality and a similar scheme with a fine-tuning step. Performance is evaluated on 5 SLU tasks from 3 dataset, namely, SLURP, FSC and SNIPS. We also used three ASR engines to investigate the impact of transcript errors and the robustness of the proposed model when golden transcripts are not available. We first show that our model can achieve comparable performance to state-of-the-art NLU models. We evaluated the robustness of our towards ASR transcripts. Results show that the proposed approach can robustly extract semantic information from audio-textual data, outperforming BERT$_{large}$ and RoBERTa$_{large}$ for low quality text transcripts from the academic CMU ASR engine. For the commercial ASR engines, we show that the MLU can be an alternative solution as it does not compromise the overall SLU performance. 

As future work, we plan to boost the results by improving the performance of our e2e SLU model. We plan to explore a low-latency MLU solution. For that, we must adapt and evaluate the proposed MLU model in a streaming setting where chunks of speech and text are synchronized and processed in an online fashion. Thus predictions of semantic labels are incrementally estimated.

\section*{Limitations}

The main limitation of this work is the performance of our e2e SLU, specially towards the more challenging SLURP dataset. As the success of our proposed MLU depends on the accuracy of both modalities involved, i.e. speech and text, guaranteeing descent performance on both modalities is important. Although we achieve competitive performance compared to the baseline results shared by the authors in \cite{bastianelli2020slurp}, the performance of our multimodal approach was more effective for low quality transcripts. Moreover, the low results of our e2e SLU corroborates with the findings in \cite{bastianelli2020slurp}, where several state-of-the-art e2e SLU were tested and were not able to surpass the proposed modular (ASR+NLU) baselines as well.

\bibliography{anthology,custom}
\bibliographystyle{acl_natbib}

\end{document}